\documentclass[journal]{IEEEtran}

\usepackage{amsmath}

\usepackage{microtype}
\usepackage{graphicx}
\usepackage{subfigure}
\usepackage{booktabs} %
\usepackage{multirow}
\usepackage{amsmath, amssymb}
\usepackage{pst-all}
\usepackage{array, arydshln}
\usepackage{paralist}
\usepackage{caption}

\newcommand{\R}{\mathbb{R}}
\newcommand{\Z}{\mathbb{Z}}

\newcommand{\doublecheckmark}{\checkmark$\!\!\!$\checkmark}

\DeclareMathOperator{\E}{E}
\DeclareMathOperator{\VAR}{VAR}

\usepackage{color}
\usepackage{cite}

\usepackage{algorithmic}
\usepackage[]{algorithm2e}

\usepackage{array}

\begin{document}

\title{Iteratively Training Look-Up Tables\\ for Network Quantization}
\author{Fabien Cardinaux, Stefan Uhlich, Kazuki Yoshiyama, Javier Alonso Garc\'{\i}a, Lukas Mauch, Stephen Tiedemann,  Thomas Kemp, Akira Nakamura%
\thanks{F. Cardinaux, S. Uhlich, K. Yoshiyama, J. Alonso Garc\'{\i}a, L. Mauch, S. Tiedemann and T. Kemp are with Sony European Technology Center, Stuttgart, Germany.}%
\thanks{A. Nakamura is with Sony Corporate, Tokyo, Japan}%
\thanks{F. Cardinaux (fabien.cardinaux@sony.com), S. Uhlich (stefan.uhlich@sony.com) and K. Yoshiyama (kazuki.yoshiyama@sony.com) are equal contributors.}%
\thanks{This work extends the preliminary study that we presented as an extended abstract at the NeurIPS 2018 CDNNRIA Workshop \cite{cardinaux2018iteratively}}.%
\thanks{\textcopyright 2019 IEEE.  Personal use of this material is permitted.  Permission from IEEE must be obtained for all other uses, in any current or future media, including reprinting/republishing this material for advertising or promotional purposes, creating new collective works, for resale or redistribution to servers or lists, or reuse of any copyrighted component of this work in other works.}}
\markboth{}%
{}
\maketitle

\begin{abstract}
Operating deep neural networks (DNNs) on devices with limited resources
requires the reduction of their memory as well as computational footprint.
Popular reduction methods are network quantization or pruning, which either reduce the word length
of the network parameters or remove weights from the network if they are not needed.
In this article we discuss a general framework for
network reduction which we call \textit{Look-Up Table Quantization} (LUT-Q).
For each layer, we learn a value dictionary and an
assignment matrix to represent the network weights.
We propose a special solver which combines gradient descent and a one-step k-means update to
learn both the value dictionaries and assignment matrices iteratively.
This method is very flexible:
by constraining the value dictionary,
many different reduction problems such
as non-uniform network quantization, training of multiplierless networks,
network pruning or simultaneous quantization and pruning can be implemented
without changing the solver. This flexibility of the LUT-Q method allows us to
use the same method to train networks for different hardware capabilities.
\end{abstract}


\section{Introduction}
\label{sec:introduction}
Deep neural networks (DNN)s are currently used in many machine learning and signal
processing applications with great success as their performance often beats the
previous state-of-the-art approaches by a large margin, e.g., see \cite{lecun2015deep}
for an overview of deep learning. DNN approaches have become
standard practice in computer vision, automatic speech
recognition and partially in natural language processing.
They are also extensively investigated to support
other domains like medicine, robotics and finance forecasting.

Recently, there has been a lot of interest in the research community in reducing the
memory/computational footprint of neural networks. This interest stems from the desire to
operate neural networks on devices with limited resources.

The most commonly used DNN reduction methods
can be categorized in the following groups of techniques:
\begin{itemize}
\item \emph{Factorized Layers} DNNs \cite{lebedev2014speeding, yunpeng2017sharing, howard2019searching,sandler2018mobilenetv2} use bottleneck architectures by factorizing traditional layers. These architecture have typically much fewer parameters than traditional DNNs.
\item \emph{Pruning} methods \cite{lecun1990handwritten, han2015learning, mauch2018least} reduce the number of weights by removing the less important connections in the DNNs.
\item \emph{Quantization} methods \cite{han2016deep, ullrich2017soft, chen2015compressing, uhlich2019dq} discretize the weights and/or activations of DNNs.
\end{itemize}

All these network reduction methods have in common
that they add structure to the weight matrices $\mathbf{W}$, which can later be used for efficient inference,
i.e., to store the weights
efficiently or to reduce the number of multiplication
and accumulation (MAC) operations.
Network pruning for example introduces (structured) sparsity, meaning
that some elements, rows or columns of $\mathbf{W}$ are set to zero and therefore can
be neglected during inference. Factorization methods introduce a low-rank structure
to $\mathbf{W}$, which can also be used for efficient inference. Quantization methods
restrict the elements of $\mathbf{W}$ to be from a restricted finite set, such that they
can be encoded with a small number of bits and thus $\mathbf{W}$ can be efficiently stored in memory.

In this article we discuss a general framework for
network reduction which we call \textit{Look-Up Table Quantization} (LUT-Q).
Primarily, LUT-Q is a non-uniform quantization method for DNNs,
which uses learned dictionaries $\mathbf{d} \in \R^K$
and lookup tables $\mathbf{A} \in \{1,...,K\}^{O \times I}$ to represent the network
weights $\mathbf{W} \in \R^{O \times I}$, i.e., we use $\mathbf{W} \in \{\mathbf{X}: [\mathbf{X}]_{oi} = [\mathbf{d}]_{[\mathbf{A}]_{oi}}, \: \mathbf{d} \in \R^K, \: \mathbf{A} \in \{1,...,K \}^{O \times I}\}$. In this article, we show that LUT-Q is a very flexible tool which allows for
an easy combination of non-uniform quantization with other reduction methods like pruning.
With LUT-Q, we can easily train networks with highly structured weight matrices $\mathbf{W}$,
by imposing constraints on the dictionary vector $\mathbf{d}$ or the assignment matrix $\mathbf{A}$.
For example, a dictionary vector $\mathbf{d}$ with $K$ elements results in quantized weights which can
be encoded with $\log_2(K) + 32K$bit.
Alternatively,
we can constrain the $\mathbf{d}$ to contain only the values $\{-1,1\}$ and
obtain a \textit{Binary Connect Network} \cite{courbariaux2015binaryconnect},
or to $\{-1,0,1\}$ resulting in a \textit{Ternary Weight Network} \cite{li2016ternary}.
This flexibility of our LUT-Q method allows us to
use the same method to train networks for different hardware capabilities. 
Moreover, we show that \mbox{LUT-Q} benefits from optimized dictionary values, compared to
 other approaches which use predefined values (e.g.~\cite{courbariaux2015binaryconnect,li2016ternary, mishra2017wrpn, mishra2018apprentice}).

The contributions of this paper are as follows:\\
\begin{compactitem}
\item We introduce LUT-Q, a trainable non-uniform quantization method
    which reduces the size and computational complexity of a DNN.
\item We propose an update rule to train DNNs which use LUT-Q.
    The update rule is a combination of a gradient descent and a k-means update,
    which can jointly learn the optimal weight dictionary $\mathbf{d}$ and assignment matrix $\mathbf{A}$.
\item We show that popular quantization methods from the literature are special cases
    of LUT-Q. By imposing specific constraints to the
    basic LUT-Q training, we can learn such networks.
\item We propose a multiplier-less
    \emph{batch normalization} (BN) that can be combined with
    LUT-Q to train \emph{fully multiplier-less} networks.
\end{compactitem}

This paper is organized as follows: Sec.~\ref{sec:related_work} summarizes
known techniques for neural network reduction and relates them
to our LUT-Q. Sec.~\ref{sec:weight_tying_networks} describes our basic
LUT-Q training algorithm and some extensions of it.
Sec.~\ref{sec:multiplierless_networks} introduces a multiplier-less
version of batch normalization which produces \emph{fully multiplier-less}
networks when combined with LUT-Q. Furthermore, Sec.~\ref{sec:hardware_considerations}
discusses efficient inference with LUT-Q networks and Sec.~\ref{sec:results} presents our experiments
and results. Finally, Sec.~\ref{sec:conclusions_future_perspectives} summarizes our
approach and gives some future perspectives of LUT-Q.

We use the following notation throughout this paper: $x$, $\mathbf{x}$, $\mathbf{X}$ and $\pmb{\mathcal{X}}$
denote a scalar, a (column) vector, a matrix and a tensor with three or four dimensions,
respectively; $\lfloor . \rfloor$ and $\lceil . \rceil$ are the floor and ceiling operators.

\section{Related Work}
\label{sec:related_work}

Different \emph{compression} methods were proposed in the past in order to reduce the memory footprint and the computational requirements of DNNs.

Common methods either reduce the number of parameters in the architecture or focus on efficient encoding of the parameters.
It has been shown that the amount of parameters of a DNN can be reduced drastically with minimal loss in performance by either designing special networks with factorized layers like in MobileNetV2 \cite{sandler2018mobilenetv2}, or alternatively
by pruning an over parametrized network \cite{lecun1990handwritten, han2015learning}.
Quantization methods allow an efficient encoding of the parameters which results in a reduced network size. LUT-Q is primarily a quantization method which can also be used effectively for pruning with only small changes in the basic algorithm.

In general, there are three categories of existing network quantization methods:
\begin{compactitem}
    \item \emph{Soft weight sharing}: These methods train the full precision weights such that they form clusters and therefore can be more efficiently quantized~\cite{nowlan1992simplifying, chen2015compressing, ullrich2017soft, louizos2017bayesian, achterhold2018variational}.

	\item \emph{Fixed quantization}: These methods choose a dictionary of values beforehand to which the weights are quantized. Afterwards, they learn the assignments of each weight to the dictionary entries. Examples are \textit{Binary Neural Networks} \cite{courbariaux2015binaryconnect}, \textit{Ternary Weight Networks} \cite{li2016ternary} and also~\cite{mishra2017wrpn, mishra2018apprentice}.

    \item \emph{Trained quantization}: These methods learn a dictionary of values to which weights are quantized during training \cite{han2016deep}.
    In \cite{han2016deep}, the authors propose to run $k$-means once after a full precision training of a DNN with float32 weights.
    As soon as the weight dictionary and assignments are obtained from the float32 network,
    they propose to fine tune the dictionary, while keeping the assignments fixed.
\end{compactitem}

The LUT-Q approach takes the best of the latter two methods: for each layer, we jointly update both dictionary and weight assignments during training. This approach to compression is similar to \emph{Deep Compression} \cite{han2016deep} in the way that we learn a dictionary and assign each weight in a layer to one of the dictionary's values. However, we run $k$-means iteratively during training and update both the assignments and the dictionary at each mini-batch iteration.

In \cite{gu2018projection} the authors introduce a quantization method that learns a set of binary weights with multiple projection matrices using backpropagation. They show several results for binary networks. %

Recently, \cite{Esser2019}, \cite{uhlich2019dq} proposed to learn the step size of a uniform quantizer by backpropagation of the training loss. While the approaches in \cite{Esser2019, uhlich2019dq} add more flexibility to fixed step size quantization, all the quantization values are constrained to be equally spaced. LUT-Q allows more flexibility by arbitrarily choosing the dictionary values.

\section{Look-Up Table Quantization Networks}
\label{sec:weight_tying_networks}

We consider training and inference of DNNs with LUT-Q layers, i.e.,
layers which compute
\begin{align}
    \mathbf{Q} &= \mathrm{LUTQ}(\mathbf{W}) \\
    \mathbf{y} &= \Phi ( \mathbf{Q} \mathbf{x} + \mathbf{b} ),
\end{align}
where $\mathbf{x} \in \R^I$ is the input vector, $\mathbf{y} \in \R^O$ is the output vector,
$\mathbf{W} \in \R^{O \times I}$ is the unquantized weight matrix, $\mathbf{Q} \in \R^{O \times I}$ is
the quantized weight matrix,
$\mathbf{b} \in \R^O$ is the bias vector and $\Phi: \R^O \rightarrow \R^O$
is the activation function of the layer.
\footnote{For simplicity, we discuss the case
where weights are represented as a matrix $\mathbf{W}$. However, LUT-Q easily extends to the
tensor case $\boldsymbol{\mathcal{W}}$ where $\boldsymbol{\mathcal{A}}$ is now also a tensor of the same size.}
$\mathrm{LUTQ}: \R^{O \times I} \rightarrow \R^{O \times I}$ is the look-up table quantization operation, which computes
\begin{equation}
    \mathrm{LUTQ}(\mathbf{W}) = \mathrm{lookup}(\mathbf{d}, \mathbf{A}),
\end{equation}
where
$\mathrm{lookup}(\mathbf{d}, \mathbf{A})$ is the table look-up operation that uses
the elements of $\mathbf{A}$ to index into the dictionary $\mathbf{d}$, i.e.,
\begin{equation}
    [\mathrm{lookup}(\mathbf{d}, \mathbf{A})]_{oi} = [\mathbf{d}]_{[\mathbf{A}]_{oi}}.
\end{equation}
At each forward pass, $\mathrm{LUTQ}(\cdot)$ first computes an optimal dictionary $\mathbf{d}  \in \R^K$ and
an assignment matrix $\mathbf{A}  \in \{1,...,K\}^{O \times I}$,
which fits best to the current weight matrix $\mathbf{W}$.
\begin{equation}
    \mathbf{d}, \mathbf{A} = \arg \min_{\mathbf{d}', \mathbf{A}'} \frac{1}{2} || \mathbf{W} - \mathrm{lookup}(\mathbf{d}', \mathbf{A}') ||^2 \\
\end{equation}
Then, the layer applies the $\mathrm{lookup}(\mathbf{d} , \mathbf{A} )$ to obtain the
quantized representation $\mathbf{Q}$ of $\mathbf{W}$ and calculates the activation $\mathbf{y}$.

\begin{figure}
\centering
		\resizebox{1.0\linewidth}{!}{
    \psset{arrowsize=2pt 3}
    \begin{pspicture}(-0.5,-0.8)(12,4)%
        \psframe(0.0,1.00)(2.0,3.00)
        \rput(1.0,0.5){\small Weights (full prec.)}
        \rput(1.0,0.15){\small $\mathbf{W} \in \R^{O\times I}$}
        \rput(1.0,-0.35){\small \emph{updated by}}
        \rput(1.0,-0.70){\small \emph{optimizer}}

        \psframe[linewidth=0.01,linestyle=dashed](0.4, 1.00)(0.6, 3.00)
        \psframe[linewidth=0.01,linestyle=dashed](0.0, 2.30)(2.0, 2.50)
        \psframe[fillstyle=hlines,hatchsep=0.025,hatchwidth=0.025,hatchcolor=gray,linewidth=0.01](0.4, 2.30)(0.6, 2.50)

        \psframe[linewidth=0.01,linestyle=dashed](1.3, 1.00)(1.5, 3.00)
        \psframe[linewidth=0.01,linestyle=dashed](0.0, 1.30)(2.0, 1.50)
        \psframe[fillstyle=hlines,hatchsep=0.025,hatchwidth=0.025,hatchcolor=gray,linewidth=0.01](1.3, 1.30)(1.5, 1.50)

        \psline[arrows=->](2.2,2.00)(3.0,2.00)
        \rput(2.6, 2.25){\small\emph{k-means}}

        \rput(3.30, 2.00){$\Bigg\{$}
        \rput(6.80, 2.00){$\Bigg\}$}

        \rput(6.1, 2.00){$,$}

        \psframe[fillcolor=white, fillstyle=solid](3.5,1.00)(5.50,3.00)
        \rput(4.40, 0.50){\small Assignments}
        \rput(4.60, 0.15){\small $\mathbf{A} \in \{1,...,K\}^{O \times I}$}

        \psframe[linewidth=0.01,linestyle=dashed](3.90, 1.00)(4.10, 3.00)
        \psframe[linewidth=0.01,linestyle=dashed](3.50, 2.30)(5.50, 2.50)

        \psframe[linewidth=0.01,linestyle=dashed](4.80, 1.00)(5.00, 3.00)
        \psframe[linewidth=0.01,linestyle=dashed](3.50, 1.30)(5.50, 1.50)

        \psframe(6.40,1.00)(6.60,3.00)
        \rput(6.60, 0.50){\small Dictionary}
        \rput(6.60, 0.15){\small $\mathbf{d} \in \R^K$}

        \psframe[fillstyle=hlines,hatchsep=0.025,hatchwidth=0.025,linewidth=0.01,linestyle=none](6.40, 1.90)(6.60, 2.10)
        \rput(6.50, 2.60){\tiny $\vdots$}
        \rput(6.50, 1.65){\tiny $\vdots$}

        \psline[arrows=->](7.00,2.00)(8.70,2.00)
        \rput(7.85, 2.25){\small\emph{Table lookup}}

        \psframe(8.90,1.00)(10.90,3.00)
        \rput(9.90,0.5){\small Weights (quantized)}
        \rput(9.90,0.15){\small $\mathbf{Q} \in \R^{O\times I}$}
        \rput(9.90,-0.35){\small \emph{used in}}
        \rput(9.90,-0.7){\small \emph{forward/backward pass}}

        \psframe[linewidth=0.01,linestyle=dashed](9.30, 1.00)(9.50, 3.00)
        \psframe[linewidth=0.01,linestyle=dashed](8.90, 2.30)(10.90, 2.50)
        \psframe[fillstyle=hlines,hatchsep=0.025,hatchwidth=0.025,linewidth=0.01](9.30, 2.30)(9.50, 2.50)

        \psframe[linewidth=0.01,linestyle=dashed](10.20, 1.00)(10.40, 3.00)
        \psframe[linewidth=0.01,linestyle=dashed](8.90, 1.30)(10.90, 1.50)
        \psframe[fillstyle=hlines,hatchsep=0.025,hatchwidth=0.025,linewidth=0.01](10.20, 1.30)(10.40, 1.50)

    \end{pspicture}}
    \caption{Proposed look-up table quantization scheme.}
    \label{fig:weight_tying:sec:weight_tying_nets}
\end{figure}
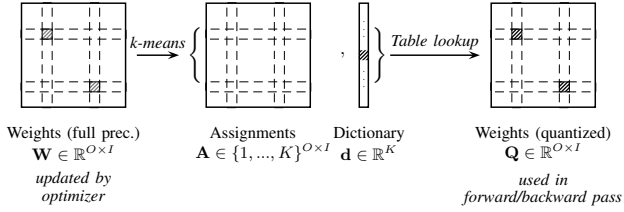

Fig.~\ref{fig:weight_tying:sec:weight_tying_nets} illustrates our LUT-Q training scheme. For each forward pass,
we learn an assignment matrix $\mathbf{A}  \in \R^{O\times I}$ and a dictionary $\mathbf{d}  \in \R^K$
and approximate the float weights $\mathbf{W}$ with $\mathbf{Q} = \mathrm{lookup}(\mathbf{d} , \mathbf{A} )$.
Hence, each weight is quantized to one of the $K$ possible values $d_1, \ldots, d_K$, according
to the indices in the assignment matrix
$\mathbf{A}$. We can use the k-means algorithm to learn $\mathbf{d}$ and assignment matrix $\mathbf{A}$.

\subsection{Training Look-Up Table Quantization Networks}
\label{sec:weight_tying_nets:subsec:training}

Training quantized neural networks with \emph{stochastic gradient descent} (SGD) is not trivial.
One problem is, that running k-means to
obtain $\mathbf{d}$ and $\mathbf{A}$ in each forward pass is
prohibitively expensive.
Furthermore, $\mathrm{LUTQ}( \cdot )$ is not differentiable, meaning that SGD can not be applied directly.
Indeed, simply quantizing the weights after each update does not work
well since the updates need to be very large (i.e., a high learning rate is required)
to change the weights to the next quantization value.
Therefore, two methodologies have been used in the literature to train networks with quantized weights:
\begin{compactitem}
    \item \emph{Incrementally quantizing weights}: after each iteration, a subset of the weights is selected and quantized. These weights are fixed for the rest of the training, while the remaining weights are still optimized \cite{zhou2017incremental}.
	\item \emph{Gradient accumulation}: the full precision weights are kept and updated during training. For the forward/backward pass the weights are quantized, i.e., gradients are computed with respect to the quantized weights; however, the update of the weights is carried out on the full precision weights. By this, information from gradients over several minibatches is accumulated and the full precision weights act as the accumulator~\cite{fiesler1990weight, courbariaux2015binaryconnect, hubara2016quantized}.
\end{compactitem}

In our work, we follow the second methodology and simply apply the straight-through gradient estimator whenever
we differentiate $\mathrm{LUTQ}(\cdot)$. Please refer to ~\cite{yin2019understanding} for an analysis of straight-through estimator. 
The gradient descent step only updates the continuous weights $\mathbf{W}$, according to
\begin{equation}
    \mathbf{W} \leftarrow \mathbf{W} - \eta \nabla_{\mathbf{W}} J(\mathbf{W}),
\end{equation}
where $J(\mathbf{W})$ is the loss function. We use the straight through gradient estimator (STE) and
simply ignore $\mathrm{LUTQ}(\cdot)$ in the backward pass, i.e., we compute
\begin{equation}
    \nabla_{\mathbf{W}} ( \mathbf{Q} \mathbf{x} + \mathbf{b} ) = \nabla_{\mathbf{W}} ( \mathbf{W} \mathbf{x} + \mathbf{b} ).
\end{equation}
Furthermore, we unroll the k-means updates over the training iterations, meaning that we just perform one
k-means update for each forward pass. This considerably reduces the computational complexity of each
forward pass, but is sufficiently accurate for training if we assume that
the continuous weights $\mathbf{W}$ do not change much between iterations (which is always the case if we use
a sufficiently small learning rate $\eta$). 

Algorithm~\ref{alg:WTN} summarizes the LUT-Q update steps for a minibatch $\{\mathbf{X},\mathbf{T}\}$, where $\mathbf{X}$
denotes the minibatch data and $\mathbf{T}$ the corresponding ground truth.
We denote the layer index by $l$ and the total number of layers by $L$.
$K^{(l)}$ is the number of values in the dictionary $\mathbf{d}^{(l)}$.
In the forward/backward pass, we use the current quantized weights $\{\mathbf{Q}^{(1)},\ldots,\mathbf{Q}^{(L)}\}$ in order to obtain the cost $C$ and the gradients $\{\mathbf{G}^{(1)},\ldots,\mathbf{G}^{(L)}\}$.
These gradients are used to update the full precision weights $\{\mathbf{W}^{(1)},\ldots,\mathbf{W}^{(L)}\}$.
Finally, using $M$ steps of $k$-means after each minibatch, we update the dictionaries $\{\mathbf{d}^{(1)},\ldots,\mathbf{d}^{(L)}\}$ and the assignment matrices $\{\mathbf{A}^{(1)},\ldots,\mathbf{A}^{(L)}\}$.
In all our experiments of Sec.~\ref{sec:results}, we use $M = 1$.
$k$-means ensures that LUT-Q is a good approximation of the full precision weights.
Note that in contrast to other approaches \cite{han2016deep}, we do not fix the assignment matrices but learn them during training.
The full precision weights $\mathbf{W}^{(l)}$ can be initialized randomly or using the weights of a previously trained full precision network.
After this initialization, for each layer, we run $k$-means in order to obtain the initial dictionary and assignment matrix\footnote{If one of the two ($\mathbf{d}$ or $\mathbf{A}$) is initially given, then we can easily compute the other one without running $k$-means.}.

In ``Step 1'', we quantize the full precision weights $\mathbf{W}^{(l)}$ to obtain $\mathbf{Q}^{(l)}$.
As described previously, we run just one $k$-means update step to update the dictionary $\mathbf{d}$ and the assignment matrix $\mathbf{A}$.
In ``Step 2'', we calculate the loss gradient with respect to the full precision weights.
In ``Step 3'' of Algorithm~\ref{alg:WTN}, we use \emph{stochastic gradient descent} (SGD)
to update the full precision weights $\mathbf{W}^{(l)}$,  with $\eta$ being the learning rate.
Please note that other optimization strategies can be used, e.g., Adam \cite{kingma2014adam}
and Nesterov accelerated gradient descent \cite{nesterov1983method}. %

\begin{algorithm}
   \caption{LUT-Q training algorithm}
   \label{alg:WTN}
\begin{algorithmic}
\begin{normalsize}
   \STATE \textcolor{gray}{// Step 1:  Compute quantized weights $LUTQ(\mathbf{W})$}
   \STATE \textcolor{gray}{// Step 1(A): Update $\mathbf{d}$ and $\mathbf{A}$ by $M$ $k$-means iterations}
   \FOR{$l=1$ {\bfseries to} $L$}
       \FOR{$m=1$ {\bfseries to} $M$}
           \STATE $A^{(l)}_{ij} = \underset{k=1,\ldots,K^{(l)}}{\arg\min} \left\lvert W^{(l)}_{ij} - d^{(l)}_k \right\rvert$
           \FOR{$k=1$ {\bfseries to} $K^{(l)}$}
           \STATE $d^{(l)}_k = \frac{1}{\sum\nolimits_{ij,\, A^{(l)}_{ij}=k} 1}\sum\nolimits_{ij,\, A^{(l)}_{ij}=k} W^{(l)}_{ij}$
           \ENDFOR
       \ENDFOR
   \ENDFOR
   \STATE \textcolor{gray}{// Step 1(B): Table lookup}
   \FOR{$l=1$ {\bfseries to} $L$}
       \STATE $\mathbf{Q}^{(l)} = \mathbf{d}^{(l)}[\mathbf{A}^{(l)}]$
   \ENDFOR

    \vspace{0.15cm}

   \STATE \textcolor{gray}{// Step 2:  Compute current cost and gradients with STE}
   \STATE $C=\text{Loss}\left(\mathbf{T},\text{Forward}\left(\mathbf{X},\mathbf{Q}^{(1)},\ldots,\mathbf{Q}^{(L)}\right)\right)$
		   \begin{align*}
		   \left\{\mathbf{G}^{(1)},\ldots,\mathbf{G}^{(L)}\right\} &= \left\{\frac{\partial C}{\partial\mathbf{Q}^{(1)}},\ldots,\frac{\partial C}{\partial\mathbf{Q}^{(L)}}\right\} \\
    &=\text{Backward}\left(\mathbf{X},\mathbf{T},\mathbf{Q}^{(1)},\ldots,\mathbf{Q}^{(L)}\right)\qquad\qquad\qquad
    		\end{align*}

   \vspace{-0.15cm}

   \STATE \textcolor{gray}{// Step 3: Update full precision weights (here: SGD)}
   \FOR{$l=1$ {\bfseries to} $L$}
       \STATE  $\mathbf{W}^{(l)} = \mathbf{W}^{(l)} - \eta \mathbf{G}^{(l)}$
   \ENDFOR

   \vspace{0.15cm}
\end{normalsize}
\end{algorithmic}
\end{algorithm}
\subsection{The flexibility of LUT-Q}
\label{sec:weight_tying_nets:subsec:examples}

While the previous section introduced the basic version of LUT-Q, we will now show that by simple modifications of the clustering (``Step 1'' of Algorithm~\ref{alg:WTN}), LUT-Q can implement many network compression schemes from the literature.

\emph{Weight pruning} reduces the network size by setting less important weights to zero \cite{reed1993pruning}.
In the LUT-Q scheme, we modify the clustering such that we impose a zero value in the first element of the dictionary, i.e.,
$\mathbf{d}^T = \begin{bmatrix} 0, \cdots \end{bmatrix}$,
and force the weights with the smallest magnitudes to be assigned to it such that we achieve a certain pruning ratio. The remaining weights are clustered using a slightly modified $k$-means where the first cluster centroid is kept to zero. I.e. $d^{(l)}_1$ in Algorithm~\ref{alg:WTN} (Step 1) is set to $0$ and not updated. Therfore all weights attracted by $d^{(l)}_1$ will be set to zero, resulting in sparse weight tensors. Han et. al.~\cite{han2016deep} show that the network size can be reduced by combining pruning with quantization. In \cite{han2016deep}, however, weights are pruned only once prior to quantization while LUT-Q allows to continuously accumulate the gradient with respect to the pruned weights (through the full precision weights) and may assign them to a non-zero value later in training.

\emph{Binary neural networks} are networks with only binary weights \cite{courbariaux2015binaryconnect} and they can be obtained with LUT-Q by choosing a fixed dictionary $\mathbf{d}^T = \begin{bmatrix} -1, +1 \end{bmatrix}$. Furthermore, if we additionally use batch normalization layers, we learn networks which are equivalent to \emph{Binary Weight Networks} \cite{rastegari2016xnor}.

\emph{Uniform quantization} can be implemented with LUT-Q by choosing and fixing the dictionary; only the weight assignments are learned. Keeping uniform quantization has computational benefits for specific hardware (e.g., fixed-point numbers).

\emph{Structured weight matrices/tensors} like Toeplitz or circulant as in \cite{sindhwani2015structured,moczulski2015acdc} are an efficient way to learn compact networks. In LUT-Q this can be done by fixing the assignment matrix to encode a specific structure and learning only the dictionary.

\emph{Multiplier-less networks} can be achieved by either choosing a dictionary $\mathbf{d}$ whose elements $d_k$ are of the form $d_k \in \{\pm2^{b_k}\}$ for all $k = 1,\ldots,K$ with $b_k \in \mathbb{Z}$, or by rounding the output of the $k$-means algorithm to powers-of-two. In this way we can learn networks whose weights are powers-of-two and can, hence, be implemented without multipliers. Multiplier-less networks will be extensively discussed in the next section.

\section{Multiplier-less Networks}
\label{sec:multiplierless_networks}
\subsection{Introduction}
\label{sec:multiplierless_networks:subsec:intro}

\begin{figure}
    \resizebox{\linewidth}{!}{\includegraphics[trim=0 0 0 30,clip]{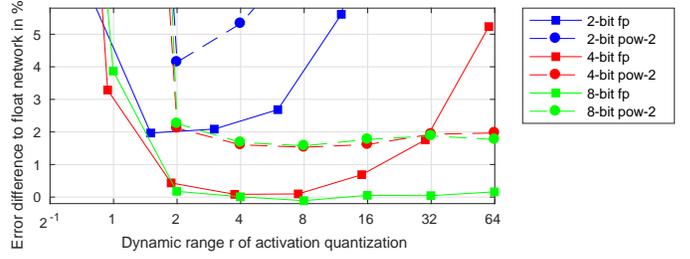}}
	\caption{CIFAR-10: Comparison of activation quantization methods (no weight quantization; $y$-axis gives validation error difference compared to network trained with float activations).}
	\label{fig:comparison_quant_methods:sec:multiplierless_networks}
\end{figure}

In general, DNN inference involves multiplications as we need to perform matrix-vector products and convolutions. Although multiplications are very optimized in modern computers, inference typically requires millions of them. Therefore, there is an interest in obtaining networks that do not require multiplications, especially for low power devices.

The idea of training multiplier-less neural networks is not new. Already in the 90's, several works \cite{white1992digineocognitron,kwan1993multiplierless,marchesi1993fast} proposed multiplier-less networks with quantization of the weights to powers-of-two and Simard and Graf \cite{simard1994backpropagation} created a multiplier-less network by using power-of-two activations. By constraining weights or activations to powers-of-two, we can drastically reduce the computational complexity of multiplications as multiplying a fixed-point number with $2^b$ becomes a bit shift by $b$ bits and multiplying a floating-point number by $2^b$ becomes an addition of $b$ to the exponent. More recently, \cite{courbariaux2016binarynet, zhou2016dorefanet, rastegari2016xnor, zhou2017incremental} have proposed approaches to train neural networks that avoid multiplications.

Power-of-two weights are easily achieved with our \mbox{LUT-Q} training. The first possibility is to set the dictionary $\mathbf{d}$ to powers-of-two (e.g., $\mathbf{d}^T = \left[\begin{smallmatrix} -1, -\frac{1}{2}, \frac{1}{2}, 1 \end{smallmatrix}\right]$  for $K = 4$) and fix it during training. A second possibility is to learn the dictionary with an additional rounding of its elements to the closest power-of-two in ``Step 1'' of the Algorithm~\ref{alg:WTN}. In order to quantize $d_k = s\cdot 2^b$ and minimize the quantization error, the quantization threshold should be the arithmetic mean between $2^{\lfloor b\rfloor}$ and $2^{\lceil b\rceil}$. Hence, the quantized value is given by
\begin{equation}
\hat d_k =
\begin{cases}
    s\cdot 2^{\lfloor b\rfloor},&  \text{if  } b - \lfloor b \rfloor \le \log_2 1.5 \\
    s\cdot 2^{\lceil b\rceil},  & \text{if  } b - \lfloor b \rfloor > \log_2 1.5
\end{cases}.
\label{eq:p2q}
\end{equation}
Both possibilities are compared in Sec.~\ref{sec:results:subsec:FQvsWT} and we will observe there that the second possibility leads to better performance.

Using power-of-two weights simplifies the calculations in affine/convolution layers. However, traditional \emph{batch normalization} (BN)~\cite{ioffe2015batch}, which has become very popular as it reduces the training time as well as the generalization gap of DNNs, still requires multiplications. Although the number of multiplications in the BN layers is typically small compared to the number of multiplications in the affine/convolution layers, removing all multiplications in a DNN is of interest for specific applications. This is the motivation for our proposed \emph{multiplier-less BN} method, which we will now describe in detail.

\subsection{Multiplier-less Batch Normalization}
\label{sec:multiplierless_networks:subsec:multiplierless_bn}
From~\cite{ioffe2015batch} we know that the traditional BN at inference time for the $o$th output is
\begin{equation}
	y_o \!=\! \gamma_o\frac{x_o - \E\left[x_o\right]}{\sqrt{\VAR\left[x_o\right]+\epsilon}} + \beta_o,
\label{eq:bn}
\end{equation}
where $\mathbf{x}$ and $\mathbf{y}$ are the input and output vectors to the BN layer, $\boldsymbol{\gamma}$ and $\boldsymbol{\beta}$ are parameters learned during training, $\E\left[\mathbf{x}\right]$ and $\VAR\left[\mathbf{x}\right]$ are the running mean and variance of the input samples, and $\epsilon$ is a small constant to avoid numerical problems. During inference, $\boldsymbol{\gamma}$, $\boldsymbol{\beta}$, $\E\left[\mathbf{x}\right]$ and $\VAR\left[\mathbf{x}\right]$ are constant and, therefore, the BN function \eqref{eq:bn} can be written as
\begin{equation}
	y_o = a_o \cdot x_o + b_o,
\label{eq:bn_simple}
\end{equation}
where we use the scale $a_o = \gamma_o / \sqrt{\VAR[x_o] + \epsilon}$ and offset $b_o = \beta_o - \gamma_o \E\left[x_o\right] / \sqrt{\VAR\left[x_o\right]+\epsilon}$. In order to obtain a multiplier-less BN, we require $\mathbf{a}$ to be a vector of powers-of-two during inference. This can be achieved by quantizing $\boldsymbol{\gamma}$ to $\boldsymbol{\hat\gamma}$. The quantized $\boldsymbol{\hat\gamma}$ is learned with the same idea as for LUT-Q: during the forward pass, we use traditional BN with the quantized $\boldsymbol{\hat\gamma} = \mathbf{\hat a} / \sqrt{\VAR[\mathbf{x}] + \epsilon}$ where $\mathbf{\hat a}$ is obtained from $\mathbf{a}$ by using the power-of-two quantization \eqref{eq:p2q}. Then, in the backward pass, we update the full precision $\boldsymbol{\gamma}$. Please note that the computations during training time are not multiplier-less but $\boldsymbol{\hat\gamma}$ is only learned such that we obtain a multiplier-less BN during inference time. This is different to \cite{courbariaux2016binarynet} which proposed a shift-based batch normalization using a different scheme that avoids all multiplications in the batch normalization operation by rounding multiplicands to powers-of-two in each forward pass. Their focus is on speeding up training by avoiding multiplications during training time, while our multiplier-less batch normalization approach avoids multiplications during inference.
\subsection{Naming Convention}
\label{sec:multiplierless_networks:subsec:naming_convention}
For the description of our results in the next sections, we will use the following naming convention:
\begin{compactitem}
    \item \emph{Quasi multiplier-less} networks avoid multiplications in all affine/convolution layers, but they are not completely multiplier-less since they contain standard BN layers, which are not multiplier-less. For example, the networks trained by Zhou et~al.~\cite{zhou2017incremental} are quasi multiplier-less.
	  \item \emph{Fully multiplier-less} networks avoid all multiplications at all. Either they are multiplier-less networks with no BN layers or multiplier-less networks that use multiplier-less BN as explained in the previous section.
    \item We call all other networks \emph{unconstrained}.
\end{compactitem}

\section{Inference Efficiency}
\label{sec:hardware_considerations}

In this section, we discuss the efficient implementation for inference of networks trained with LUT-Q. %
Note that our objective is to reduce the memory footprint, required computations and energy consumption of neural networks at inference time. %
We start by briefly discussing the reduction that can be achieved in terms of memory footprint and then discuss the reduction in the number of computations.

The memory used for network parameters is dominated by the weights in affine/convolution layers. Using LUT-Q, instead of storing $\mathbf{W}$, the dictionary $\mathbf{d}$ and the assignment matrix $\mathbf{A}$ are stored. Hence, for an affine/convolution layer with $N$ parameters, the reduction is
\begin{equation}
    N B_\text{float} \xrightarrow{LUT-Q} KB_\text{float} + N\left\lceil \log_2 K\right\rceil,
    \label{eq:weight_tying_size}
\end{equation}
where $B_\text{float}$ is the number of bits to store a weight (e.g., $B_\text{float} = 32$ bit for single precision). %
In Sec.~\ref{sec:results}, we experiment with $K$ ranging from $K=2$ (1-bit quantization) to $K=256$ (8-bit quantization).

Assuming that the device used for inference performs a forward pass by computing layer operations sequentially, the device typically keeps a buffer to store the input and output activations of the layer being processed. Under this assumption, the buffer memory should be large enough to store the input and output activations of any layer in the network. Table~\ref{tab:wt_footprint} shows that for full precision networks, the buffer memory is one order of magnitude smaller than the memory needed for the parameters. However, for heavily quantized networks the buffer memory used for the activations dominates the memory footprint and, therefore, it should be quantized as well. Fig.~\ref{fig:comparison_quant_methods:sec:multiplierless_networks} compares ``fp'' and ``pow-2'' activation quantization.\footnote{Please refer to Sec.~\ref{sec:results:subsec:FQvsWT} for more details about fp and pow-2 quantization. In contrast to weight quantization, we only need to quantize to $[0, r]$ due to the ReLU nonlinearity, i.e., we do not need to spend a bit on coding the sign.} This plot shows that uniformly quantizing the activations to 8-bit with a well chosen dynamic range $[0, r]$ allows to reduce the required buffer memory by a factor of four without loss in accuracy. We will use this 8-bit activation quantization for the remaining experiments presented in this paper except where explicitly stated.

Using LUT-Q or other quantization methods, we also achieve a reduction in the number of computations. Consider the case of computing one output value $y_o$ for a LUT-Q affine layer which is given by
\begin{equation}
    y_o = b_o + \sum_{i = 1}^I Q_{oi} x_i = b_o + \sum_{k = 1}^K d_k \left( \sum_{i=1,\, A_{oi} = k}^I x_i\right),
    \label{eq:weight_tying_comp}
\end{equation}
where $I$ denotes the size of the input vector $\mathbf{x}$. We reduce the number of multiplications from $I$ to $K$. Similarly, in the case of a 2-D convolution layer we reduce the multiplications for one output map from $I \cdot S \cdot F$ to $S \cdot K$, where $I$ is now the number of input maps, $S$ is the map size (height $\times$ width) and $F$ is the filter size (height $\times$ width). The efficient hardware implementation in \eqref{eq:weight_tying_comp} is achieved by $K$ parallel registers that store the sum of activations for each $k$.

Table~\ref{tab:wt_footprint} summarizes the memory footprint and computations for the image classification networks that we will use in the next section.

\begin{table}
\caption{Memory and computations for ResNet-20 for CIFAR-10 and ResNet-18/-34/-50 for ImageNet. Activations have 32 bit.}
\label{tab:wt_footprint}
\begin{scriptsize}
\centering
\renewcommand{\arraystretch}{1}
\renewcommand{\tabcolsep}{6pt}
\resizebox{\linewidth}{!}{%
\begin{tabular}{crrrrr}
    \toprule[1.5pt]
		\multirow{2}{*}{\textbf{Net}}   & \textbf{Weight}    & \textbf{Param.} & \textbf{Buffer} & \multicolumn{2}{c}{\textbf{Computations}} \\
		                                & \textbf{Quant.}  & \textbf{Memory} & \textbf{Memory} & \multicolumn{2}{c}{\textbf{(million of ops)}} \\
																		&  									 & \textbf{(MB)}   & \textbf{(MB)}   & \textbf{Add.} & \textbf{Mul.} \\
    \midrule[1.5pt]
    \multicolumn{6}{c}{\emph{\textbf{ResNet-20 for CIFAR-10}}} \\
		Full Prec.   & $32$-bit & $1.03$ & $0.13$  & $40.64$ & $40.55$ \\
		LUT-Q & $8$-bit & $0.28$ & $0.13$ & $40.64$ & $32.56$ \\
		LUT-Q & $4$-bit & $0.13$ & $0.13$ & $40.64$ & $3.01$ \\
		LUT-Q & $2$-bit & $0.07$ & $0.13$ & $40.64$ & $0.75$ \\
		LUT-Q & $1$-bit & $0.04$ & $0.13$ & $40.64$ & $0.38$ \\
    \midrule
    \multicolumn{6}{c}{\emph{\textbf{ResNet-18 for ImageNet}}} \\
        Full Prec.  & $32$-bit & $44.59$ & $3.64$ & $1814.85$ & $1814.07$  \\
        LUT-Q & $4$-bit & $5.61$ & $3.64$ & $1814.85$ & $39.76$  \\
		LUT-Q & $2$ bit & $2.83$ & $3.64$ & $1814.85$ & $9.94$  \\
	\midrule
	\multicolumn{6}{c}{\emph{\textbf{ResNet-34 for ImageNet}}} \\
		Full Prec.  & $32$-bit & $83.15$ & $3.64$ & $3665.17$ & $3663.76$ \\
		LUT-Q & $4$-bit & $10.46$ & $3.64$ & $3665.17$ & $59.83$  \\
		LUT-Q & $2$-bit & $5.26$ & $3.64$ &  $3665.17$ & $14.96$  \\
    \midrule
    \multicolumn{6}{c}{\emph{\textbf{ResNet-50 for ImageNet}}} \\
		Full Prec.  & $32$-bit & $97.49$ & $4.59$ & $4094.80$ & $4089.18$ \\
		LUT-Q & $4$-bit & $12.37$ & $4.59$ & $4094.80$ & $177.84$  \\
		LUT-Q & $2$-bit & $6.29$ & $4.59$ & $4094.80$ & $44.46$ \\
    \bottomrule[1.5pt]
\end{tabular}}
\end{scriptsize}
\end{table}

\section{Experiments}
\label{sec:results}

We conducted extensive experiments with LUT-Q quantization on the CIFAR-10 image classification benchmark \cite{krizhevsky2009learning}. We first confirm the potential of learning a dictionary of quantized values. Then we demonstrate the capability of LUT-Q to train pruned and multiplier-less networks. Then, we evaluate LUT-Q on large scale tasks, namely the ImageNet ILSVRC-2012 image classification~\cite{russakovsky2015imagenet}, Pacscal VOC object detection~\cite{everingham2009ThePV} and the Wall Street Journal acoustic modeling~\cite{paul1992design} for \emph{automatic speech recognition} (ASR).

All experiments are implemented, using the \textit{Sony Neural Network Library}\footnote{Neural Network Libraries by Sony: https://nnabla.org/}. We implemented efficiently the k-means updates in CUDA, using monolithic kernels, vectorized loads operations and shuffle instructions. This reduces drastically the training time overhead due to the k-means updates.

For CIFAR-10, we first trained the full precision \textit{ResNet-20}, which is used to initialize the quantization trainings (seed network). We followed the training procedure used in~\cite{he2016deep} with data augmentation: we trained for $160$ epochs, decreasing the learning rate at epochs $80$ and $120$. For training the quantized network we started from the seed network and applied quantization training (e.g., LUT-Q) for additional $160$ epochs, following the same training scheme. Performance is evaluated as the best validation error without image augmentation on the validation data.

The seed network achieves an error rate of $7.84\%$. However, for a fair comparison to the quantized networks, we continued training it for additional $160$ epochs and achieved the slightly lower error rate of $7.42\%$ (averaged over $10$ runs). We use this error rate as our 32-bit full precision network baseline.

\subsection{Fixed Quantization versus LUT-Q}
\label{sec:results:subsec:FQvsWT}
One of the main advantages of LUT-Q is that it jointly learns both dictionary values and assignments. Traditional approaches fix the quantization values in advance and just learn the assignments. Common choices for the quantization to $n$ bits are \emph{fixed-point} or \emph{powers-of-two}:
\begin{compactitem}
	\item \emph{Fixed-point quantization} (``fp'') quantizes a float weight $w = s\lvert w\rvert$ to
    \begin{equation*}
        q = s\cdot\delta\cdot\begin{cases}\left\lfloor \frac{\lvert w\rvert}{\delta} + 0.5\right\rfloor  & \lvert w \rvert / \delta \le \left(2^{n-1} - 1\right) \\ 2^{n-1} - 1 & \text{otherwise} \end{cases},
    \end{equation*}
    i.e., uniformly quantizes $w$ using the quantization step size $\delta$ and, hence, the dynamic range is $r = (2^{n-1} - 1)\delta$. For an efficient hardware implementation, $\delta$ needs to be a power-of-two as the multiplication of two fp quantized numbers can then be implemented by an integer multiplication and a bit shift by $n$ bits.

    \item \emph{Pow-2 quantization} (``pow-2''): Similarly, the pow-2 quantization of the weight $w = s\lvert w\rvert$ is given by
    \begin{equation*}
        q = s\cdot\begin{cases}
            0 & \lvert w \rvert \le 2^{m - 2^{n-2} + 0.5}\\
            2^{\left\lfloor \log_2 \lvert w \rvert + 0.5\right\rfloor}  & 2^{m - 2^{n-2} + 0.5} < \lvert w \rvert \le 2^m \\
            2^m & \text{otherwise} \end{cases},
    \end{equation*}
    where $m \in \Z$ with $2^m = r$ being the dynamic range.
\end{compactitem}

We use the same scheme as~\cite{mishra2017wrpn} and~\cite{mishra2018apprentice} to train these networks. Note that this is equivalent to LUT-Q training but skipping the dictionary update in the initial $k$-means and in the ``Step 1'' of Algorithm~\ref{alg:WTN}.

For fp and pow-2 quantization, we need to choose the dynamic range $[-r, r]$ for the weight quantization. Preliminary experiments showed that the choice of the dynamic range drastically influences the performance, especially for very small bitwidth quantization. We follow the approach from~\cite{zhou2017incremental} where the dynamic range is chosen for each layer using $r = 2^{\left\lceil \log_2 \max_{oi} \lvert W_{oi} \rvert \right\rceil}$. As explained in Sec.~\ref{sec:hardware_considerations}, full precision activations dominate the memory requirements for very low bitwidth weights. Therefore, we trained the networks with activations quantized to 8-bit using fp quantization.

Table~\ref{tab:CIFAR-quantized} compares the results of 4-bit and 2-bit quantization of a ResNet-20 with either fixed point (fp) or LUT-Q (constrained to power-of-two). We observe that fp achieves near baseline performances, however, we lose performance when the weights are quantized to 2 bit. LUT-Q achieves similar performance for 4-bit networks and significantly outperformed the fixed quantization methods for 2-bit weights with $8.02\%$ error rate, even if we constrain the values to be powers-of-two.

\begin{table}[t]
\caption{CIFAR-10: Validation error of ResNet-20 with 4-bit and 2-bit weight quantization. The full precision baseline model achieves an error rate of $7.42\%$.}
\label{tab:CIFAR-quantized}
\centering
\normalsize
\renewcommand{\tabcolsep}{2pt}
\resizebox{\linewidth}{!}{
\begin{tabular}{lccccrr}
\toprule
\textbf{Weight quant.} & \textbf{Batch norm}   & \multicolumn{2}{c}{\textbf{Activation quant.}}   && \multicolumn{2}{c}{\textbf{Validation error}}\\
 method & method & bitwidth & method && \multicolumn{1}{c}{4-bit} & \multicolumn{1}{c}{2-bit}\\
\midrule
fp            & traditional     & 8-bit & fp       &&  \bf{7.60\%}  & 12.72\% \\
pow-2 LUT-Q   & traditional     & 8-bit & pow-2    &&  7.61\%  & \bf{8.02\%}  \\
\bottomrule
\end{tabular}}
\end{table}

\subsection{Pruning}
\begin{figure}[t]
    \centering
    \resizebox{0.7\linewidth}{!}{\includegraphics[trim=0 0 0 0,clip]{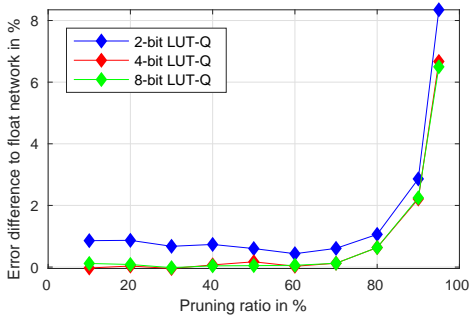}}
    \caption{CIFAR-10: Validation error for LUT-Q with pruning.}
    \label{fig:cifar10_pruning}
\end{figure}
As explained in Sec.~\ref{sec:weight_tying_nets:subsec:examples}, LUT-Q can be used to prune and quantize networks. As seen in Sec~\ref{sec:hardware_considerations}, quantization is a very effective way of reducing the memory size and also the number of multiplications of a network. From Table~\ref{tab:wt_footprint}, we observe that the remaining computations are dominated by the additions. Interestingly, from \eqref{eq:weight_tying_comp}, when using LUT-Q for pruning and quantization, $d_1=0$ and all additions for $k=1$ are avoided. Therefore, the reduction in number of additions in affine/convolution layers is proportional to the pruning ratio.

Figure~\ref{fig:cifar10_pruning} shows the error rate increase between the baseline full precision ResNet-20 and the pruned and quantized network using full precision activations. Using LUT-Q we can prune and quantize the networks up to $70\%$ without significant loss in performance. With this pruning ratio we reduce the total number of additions from $40.64$M to $12.38$M.

Thanks to the ability of LUT-Q to simultaneously prune and quantize networks, we have shown that we can drastically reduce the memory footprint, the number of multiplications and the number of additions of deep neural networks at the same time.

\subsection{Multiplier-less Networks}

While the number of multiplications can be reduced using standard LUT-Q, remaining multiplications in affine/convolution layers can be avoided by restricting the dictionary to powers-of-two. Sec.~\ref{sec:multiplierless_networks} described how to train such a network with LUT-Q. We refer to this method as \textit{pow-2 LUT-Q}. In our experiments, constraining the dictionary to powers-of-two did not degrade the performance compared to an unconstrained dictionary. Therefore, in the remaining of this paper we will not show unconstrained LUT-Q results and will only focus on pow-2 LUT-Q.

Recently, Zhou et~al.~\cite{zhou2017incremental} proposed an \emph{incremental network quantization} (INQ) approach to train quasi multiplier-less\footnote{The authors refer to them as multiplier-less but, as explained in Sec.~\ref{sec:multiplierless_networks:subsec:multiplierless_bn}, multiplications are still required for the batch normalization layers. We call these networks \textit{quasi multiplier-less}, see Sec.~\ref{sec:multiplierless_networks:subsec:naming_convention}.} networks. Since Zhou et~al.~\cite{zhou2017incremental} did not conduct experiments on the ResNet-20 dataset, we compare LUT-Q with our own implementation of INQ in Table~\ref{tab:CIFAR-multiplierless-table}. \footnote{Note that INQ networks are trained with the native INQ implementation from \textit{Sony Neural Network Library}} For  training we initialized ResNet-20 with the same full precision model and trained the INQ network for the same number of epochs (i.e., $160$ epochs) as for LUT-Q. Every $20$ epochs we quantized and fixed half of the weights based on the so-called \textit{pruning-inspired partition} criterion. We found that we obtained best results when we reduced the learning rate twice within each period of $20$ epochs and reset it to the original learning rate after quantizing and fixing more weights. Table~\ref{tab:CIFAR-multiplierless-table} compares the performance of the LUT-Q networks to the performance of quasi multiplier-less networks quantized to $\left\lbrace8,4,2\right\rbrace$ bits with INQ.
We can observe, that LUT-Q systematically outperform INQ for both quasi and fully multiplier-less networks.
This can be best observed for networks which use a very small bitwidth.
Second, to construct fully multiplier-less networks, it is beneficial to choose a network with power-of-two quantized weights which uses a multiplier-less batch normalization. Such networks consistently outperform multiplier-less networks with
power-of-two activations and traditional batch normalization. Moreover, we observe that  the best fully multiplier-less network lose around $0.5$\% to $2$\% accuracy compared to the quasi multiplier-less networks as usually reported in the literature. 

Most of the multiplications are typically in the affine/convolution layers. However, for networks like ResNet, some multiplications remain in the batch normalization layers during inference. To get rid of these multiplications, we can think of three approaches:
\begin{compactitem}
\item Remove batch normalization layers and train without them.
\item Quantize the activations to powers-of-two: In this case, it is not required to enforce power-of-two weights and traditional batch normalization can be used, since the coefficients of the batch normalization can be folded into the weights from a preceding affine/convolution layer during inference.
\item Replace traditional batch normalization by a multiplier-less batch normalization as described in Sec.~\ref{sec:multiplierless_networks:subsec:multiplierless_bn}.
\end{compactitem}
In our experiments, training quantized networks without batch normalization layers is difficult and leads to poor performance. For example, the error rate for 8-bit weight quantization is more than $5\%$ higher than for the full precision baseline with batch normalization. Table~\ref{tab:CIFAR-multiplierless-table} shows the results of quasi multiplier-less against fully multiplier-less networks. We observe some performance loss when constraining to fully multiplier-less networks. However, the models with multiplier-less batch normalization systematically outperform the models with power-of-two activations. Furthermore, the networks trained with LUT-Q outperform those trained with INQ.

For 4-bit power-of-two weights and 8-bit activations, the error rate of LUT-Q quasi multiplier-less ResNet-20 is only $0.19\%$ higher than the baseline full precision model and the fully multiplier-less version has an increased error of $0.65\%$ compared to the baseline.

\begin{table*}[t]
\caption{CIFAR-10: Validation error of multiplier-less ResNet-20 with 8-bit/4-bit/2-bit/1-bit weight quantization. The full precision baseline model achieves an error rate of $7.42\%$.}
\label{tab:CIFAR-multiplierless-table}
\centering
\normalsize
\renewcommand{\tabcolsep}{2pt}
\begin{tabular}{lccccrrrr}
\toprule
\textbf{Weight quant.} & \textbf{Batch norm}   & \multicolumn{2}{c}{\textbf{Activation quant.}}   && \multicolumn{4}{c}{\textbf{Validation error}}\\
 method & method & bitwidth & method && \multicolumn{1}{c}{8-bit} & \multicolumn{1}{c}{4-bit} & \multicolumn{1}{c}{2-bit} & \multicolumn{1}{c}{1-bit}\\
\midrule
\multicolumn{9}{c}{\emph{\textbf{Quasi multiplier-less}}} \\
INQ      & traditional     & 8-bit & fp  && 8.34\%  & 8.46\%  & 38.84\% & \multicolumn{1}{c}{ - }     \\
pow-2 LUT-Q & traditional     & 8-bit & fp  && \textbf{7.22\%}  & \textbf{7.61\%}  & \textbf{8.02\%}  &  \textbf{9.31\%} \\
\midrule
\multicolumn{9}{c}{\emph{\textbf{Fully multiplier-less}}} \\
INQ      & multiplier-less & 8-bit & fp  && 10.61\% & 10.58\% & 39.47\% & \multicolumn{1}{c}{ - }     \\
LUT-Q       & traditional     & 8-bit & pow-2    && 8.72\%  & 8.99\%  & 10.37\% & 13.82\% \\
pow-2 LUT-Q & multiplier-less & 8-bit & fp  && 8.23\%  & 8.07\%  & 8.98\%  & 11.58\% \\
\bottomrule
\end{tabular}%
\end{table*}

\subsection{ImageNet Experiments}
\begin{table*}
\caption{\small ImageNet: Multiplier-less networks using LUT-Q. \doublecheckmark:~fully multiplier-less. \checkmark:~quasi multiplier-less. $\times$:~unconstrained.}
\label{tab:IMAGENET-multiplierless-table}
\centering
\normalsize
\begin{tabular}{rllllclcrccl}
\toprule
\multicolumn{2}{c}{\multirow{2}{*}{\textbf{Weights}}} && \multirow{2}{*}{\textbf{Batch norm}} & & \multicolumn{2}{c}{\multirow{2}{*}{\textbf{Activations}}} & \textbf{Multiplier-} & \multicolumn{3}{c}{\textbf{Validation error}} \\
& & & & &  & & \textbf{less} &  ResNet-18 & ResNet-34 & ResNet-50  \\
\hline
32-bit &          && traditional     && 32-bit & & $\times$  &  30.96\% & 28.06\% & 25.87\% \\
\hline
4-bit  & LUT-Q       && traditional     && 8-bit  & pow-2 & \doublecheckmark & 34.37\% & 31.44\% & 27.50\% \\
4-bit  & LUT-Q pow-2 && multiplier-less && 8-bit & fp    & \doublecheckmark & 35.06\% & 30.65\% & 26.89\% \\
4-bit  & LUT-Q pow-2 && traditional     && 8-bit & fp    & \checkmark & 31.58\% & 28.10\% & 25.46\%\\
\hline
2-bit  & LUT-Q       && traditional     && 8-bit  & pow-2 & \doublecheckmark & 43.16\% & 37.34\% & 32.12\% \\
2-bit  & LUT-Q pow-2 && multiplier-less && 8-bit & fp    & \doublecheckmark & 43.23\% & 35.20\% & 29.81\%  \\
2-bit  & LUT-Q pow-2 && traditional     && 8-bit & fp    & \checkmark & 35.80\% & 30.51\% & 26.92\% \\
\bottomrule
\end{tabular}%
\end{table*}

\begin{table*}
\caption{ImageNet: LUT-Q compared to other quantization methods. \checkmark:~quasi multiplier-less. $\times$:~unconstrained. The HAQ \cite{haq2019} results are for networks of the same size as 2-bit and 4-bit quantizated networks but with variable bitwidth per layer.}
\label{tab:IMAGENET-compared}
\normalsize
\setlength\extrarowheight{2pt}
\setlength\tabcolsep{1.5pt}
\begin{center}
\begin{tabular}{rlrllcccc}
\toprule
\multicolumn{3}{c}{\textbf{Quantization}}     &&  \multirow{2}{*}{\textbf{Source}} & \textbf{Mulitplier-} & \multicolumn{3}{c}{\textbf{Validation error}}\\
\multicolumn{2}{c}{Weights} & Activations &&  	                              &  \textbf{less}                       &  ResNet-18 & ResNet-34 & ResNet-50\\
\hline
32-bit &      & 32-bit    && our implementation             & $\times$  & 31.0\% & 28.1\% & 25.9\% \\
\hline
5-bit & pow-2 & 32-bit   && INQ \cite{zhou2017incremental} & \checkmark & 31.0\% & -      & 25.2\% \\
\hline
4-bit & pow-2 & 32-bit   && INQ \cite{zhou2017incremental} & \checkmark & 31.1\% & -      & -      \\
$\thicksim$4-bit  && 32-bit   && HAQ \cite{haq2019} & $\times$  & -  & -      & 24.9\%      \\
\hline
4-bit  &      & 8-bit    &&  Mishra \it{et. al} \cite{mishra2018apprentice}   & $\times$   & 33.6\% & 29.7\% & 28.5\% \\
4-bit & pow-2 & 8-bit    &&  \textbf{LUT-Q pow-2} (ours)       & \checkmark & 31.6\% & 28.1\% & 25.5\% \\
\hline
2-bit & pow-2 & 32-bit   && INQ \cite{zhou2017incremental} & \checkmark & 34.0\% & -      & -      \\
2-bit &       & 32-bit   &&  Mishra \it{et. al} \cite{mishra2018apprentice}    & $\times$   & 33.4\% & 28.3\% & 26.1\% \\
2-bit & pow-2 & 32-bit    && \textbf{LUT-Q pow-2} (ours)      & \checkmark & 31.8\% & -      & -      \\
$\thicksim$2-bit  && 32-bit   && HAQ \cite{haq2019} & $\times$  & -  & -      & 29.4\%      \\
\hline
2-bit   &     & 8-bit    &&  Mishra \it{et. al} \cite{mishra2018apprentice}    & $\times$   & 33.9\% & 30.8\% & 29.2\% \\
2-bit & pow-2 & 8-bit    && \textbf{LUT-Q pow-2}  (ours)      & \checkmark & 35.8\% & 30.5\% & 26.9\% \\
\bottomrule
\end{tabular}
\end{center}
\end{table*}

\begin{figure*}[t]
\centering
\resizebox{0.8\linewidth}{!}{\includegraphics[trim=25 6 17 18,clip]{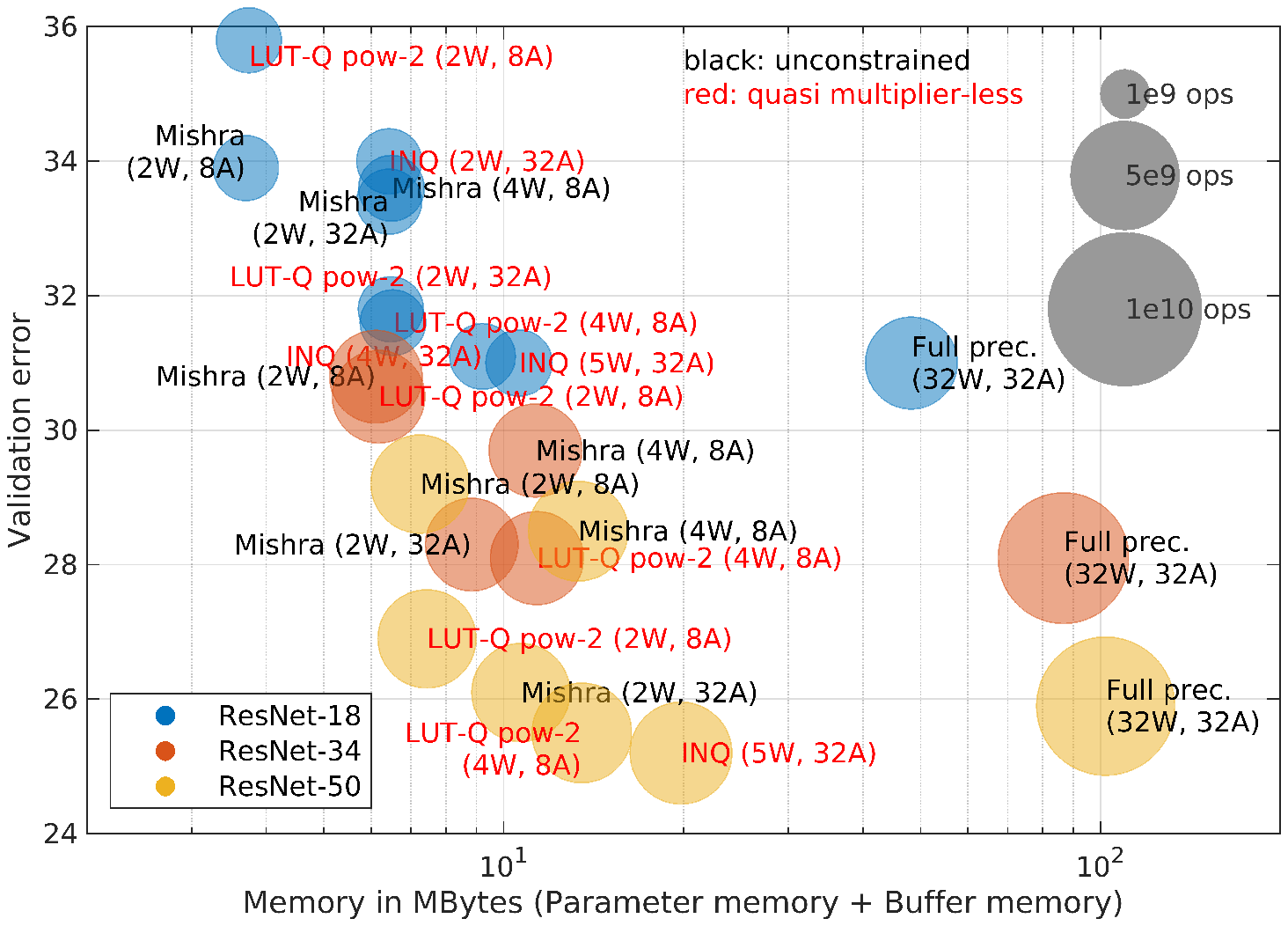}}
\caption{ImageNet: LUT-Q graphically compared to other quantization methods: the bubble size gives number of operations, i.e., the sum of \#additions and \#multiplications; ``$n_1$W, $n_2$A'' refers to $n_1$-bit weights and $n_2$-bit activations.}
\label{fig:IMAGENET-compared}
\end{figure*}

We also trained quantized and multiplier-less networks on the more challenging ImageNet task. We use \textit{ResNet-18}, \textit{ResNet-34} and \textit{ResNet-50} as reference networks \cite{he2016deep}. For training the ResNet models on ImageNet, we first resized all images to $320 \times 320$, then applied data augmentation (aspect ratio, flipping, rotation, etc.) and finally randomly cropped to a $224 \times 224$ image. We train the models for $90$ epochs with SGD using Nesterov momentum starting with a learning rate of $0.1$ and decay it by a factor of $10$ every $30$ epochs. We followed the same training scheme for the quantized networks and initialized their parameters with the trained full precision parameters. Reported performance are top-1 errors of the center-cropped validation images.

Results of multiplier-less LUT-Q training for ResNet-18, ResNet-34 and ResNet-50 are shown in Table~\ref{tab:IMAGENET-multiplierless-table}. Note that all the quantized networks trained with LUT-Q also use activation quantization to 8 bit, which avoids some of the overhead due to the buffer memory as explained in Sec.~\ref{sec:hardware_considerations}.

For fully multiplier-less networks, the models with multiplier-less batch normalization outperform quantizing the activations to powers-of-two for the two larger ResNets (ResNet-34 and Resnet-50). However, for ResNet-18, quantizing the activations to powers-of-two leads to slightly better results than using multiplier-less batch normalization.

Better performance is achieved at the cost of keeping the multiplications in the batch normalization. Remarkably, ResNet-50 with 2-bit weights and 8-bit activations achieves $26.92\%$ error rate which is only $1.05\%$ worse than baseline. This memory footprint of this network (for parameters and activations) is only $7.35$MB compared to $97.5$MB for the full precision network. Furthermore, the number of multiplications is reduced by two orders of magnitude and most of them can even be replaced by bit-shifts.

We are not aware of any other published result with both power-of-two weights and quantized activations. In Table~\ref{tab:IMAGENET-compared}, we compare LUT-Q against the following methods reported in the literature:
\begin{itemize}
\item The INQ approach, which also trains networks with power-of-two weights.
\item The best results from literature with 4-bit or 2-bit weight quantization and full precision or 8-bit activation quantization, collected by~\cite{mishra2018apprentice}.
\item The HAQ approach recently reported in~\cite{haq2019} which achieves state-of-the-art performance for ResNet with quantized weights. The reported HAQ results are obtained with a method similar to deep compression \cite{han2016deep} and by learning the optimal bitwidth for each layer in order to achieve the same network memory footprint than 2-bit and 4-bit networks. The activations, however, are not quantized for the reported HAQ results.
\end{itemize}

Note that we cannot directly compare the results from the \textit{apprentice} method proposed in~\cite{mishra2018apprentice} and \textit{LSQ}~\cite{Esser2019} because they do not quantize the first and last layers of the ResNets. We observe that LUT-Q always achieves better performance than other methods with the same weight and activation bitwidth except for ResNet-18 with 2-bit weight and 8-bit activation quantization. Additionally, \mbox{LUT-Q} networks are superior to the other networks in the sense that they combine activation quantization and power-of-two weights, i.e., most multiplications can be replaced by simpler bit-shifts.

\subsection{Object Detection Experiments}
We evaluated the performance of LUT-Q for object detection on the Pascal VOC \cite{everingham2009ThePV} dataset. We use our implementation of YOLOv2 \cite{redmon2017yolo9000} as baseline. This network has a memory footprint of 200MB and achieves a \textit{mean average precision (mAP)} of $72\%$ on Pascal VOC. We were able to reduce the total memory footprint by a factor of 20 while maintaining the mAP above $70\%$ by carrying out several modifications: replacing the feature extraction network with traditional residual networks \cite{he2016deep}, replacing the convolution layers by factorized convolutions \footnote{Each convolution is replaced by a sequence of pointwise, depthwise and pointwise convolutions (similarly to MobileNetV2 \cite{sandler2018mobilenetv2}).}, and finally applying \mbox{LUT-Q} in order to quantize the weights and activations of the network to 8 bit. Note that the feature extractor with residual and factorized layers was pre-trained on the ImageNet dataset, however, the \mbox{LUT-Q} quantization training was performed directly on the Pascal VOC object detection task.

\begin{figure}
    \resizebox{\linewidth}{!}{\includegraphics[trim=0 0 0 0,clip]{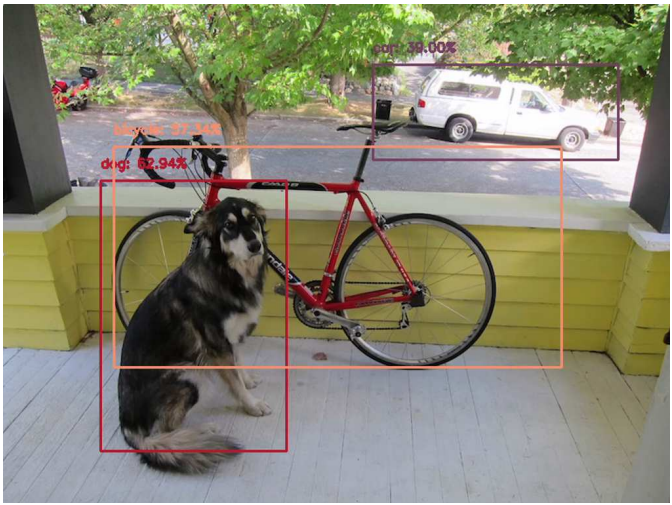}}
	\caption{Example of object detection results with a very low memory footprint DNN. Weights and activations are quantized to 4 bit. The total memory footprint is only 1.72MB (more than 100 times smaller than YOLOv2 \cite{redmon2017yolo9000}).}
	\label{fig:yolo_example}
\end{figure}

When we further quantize weights and activations to 4 bit, we are able to reduce the total memory footprint down to just 1.72MB and still achieve a mAP of about $64\%$. An example of the object detection results for this 4-bit quantized network can be seen on Fig.~\ref{fig:yolo_example}.

\subsection{Automatic Speech Recognition Experiments}

In this section, we report our evaluation of LUT-Q on an acoustic model for \emph{automatic speech recognition} (ASR). We use a fully connected network with wide layers, in contrast to convolution neural networks used in the previous image classification tasks, which typically have a smaller number of parameters per layer. Moreover, acoustic models are trained on very large datasets, typically one or two orders of magnitude larger than ImageNet. The ASR dataset used in these experiments has $20$ times more samples than the ImageNet dataset.

For these experiments we use features extracted from the \emph{Wall Street Journal} (WSJ) speech corpus~\cite{paul1992design}; the feature vectors have a size of $440$. The task is a large classification task with $3407$ different classes. The network output provides the \textit{senone} probabilities that are used later by the \emph{decoder} in order to recognize a speech sequence. The WSJ training dataset contains $26.5$ million samples and the WSJ validation set contains $2.65$ million samples. The used DNN model has seven wide fully connected layers with ReLU activations, the output of the network is a \textit{Softmax} layer. This model does not use Batch Normalization and does not have any Dropout layers.

The full precision network achieves $35.1\%$ validation error and is used to initialize the quantization training. As the large layer size slows down the clustering in ``Step 1'' of Algorithm~\ref{alg:WTN}, we run the update of $\mathbf{A}^{(l)}$ and $\mathbf{d}^{(l)}$ only every $50$ minibatches. For this task, activations were not quantized as their memory footprint remains small compared to the memory footprint of the weights for the large affine layers.

Table~\ref{tab:asr_wt} shows the performance of the LUT-Q quantization. These results confirm that LUT-Q can be successfully applied for different architectures -- in this example for a network with wide affine layers.

\begin{table}
\caption{WSJ: Comparison of multiplier-less LUT-Q networks for acoustic modeling.}
\label{tab:asr_wt}
\vskip 0.15in
\centering
\begin{small}
\centering
\renewcommand{\arraystretch}{1}
\renewcommand{\tabcolsep}{4pt}
\begin{tabular}{llrr}
    \toprule
		\textbf{Method}   & \textbf{bitwidth}  & \textbf{Param. Memory}  & \textbf{Val. Error}\\
		\midrule
		Full Precision   & $32$-bit        & 923.7 Mbit & $35.1$\% \\
		\midrule
	    LUT-Q pow2  & $4$-bit         & 115.9 Mbit & $35.6$\% \\
	    LUT-Q pow2   & $3$-bit         &  87.1 Mbit & $36.4$\% \\
		LUT-Q pow2 & $2$-bit 			  &  58.2 Mbit & $36.8$\% \\
		LUT-Q pow2 & $1$-bit (binary) &  29.3 Mbit & $38.6$\% \\
		\bottomrule
\end{tabular}
\end{small}
\end{table}

\section{Conclusions and Future Perspectives}
\label{sec:conclusions_future_perspectives}

We have presented LUT-Q, a novel approach for the reduction of size and computations for deep neural networks. After each minibatch update, the quantization values and assignments are also updated by a clustering step. We show that our LUT-Q approach can be efficiently used for pruning weight matrices and training multiplier-less networks as well. We also introduce a new form of batch normalization that avoids the need for multiplications during inference.

As argued in this paper, if weights are quantized to very low bitwidth, the activations may dominate the memory footprint of the network during inference. Therefore, we perform our experiments with activations uniformly quantized to 8 bit. We believe that a non-uniform activation quantization, where the quantization values are learned parameters, will help quantize activations to lower precision. This is one of the promising directions for continuing this work.

Recently, several papers have shown the benefits of training quantized networks using a \textit{distillation} strategy \cite{hinton2015distilling,mishra2018apprentice}. Distillation is compatible with our training approach and we are planning to investigate LUT-Q training with distillation.

\bibliographystyle{IEEEtran}
\bibliography{papers}

\end{document}